\documentclass{article}

\usepackage{microtype}
\usepackage{graphicx}

\usepackage{caption}
\usepackage{subcaption}
\usepackage{booktabs} %

\usepackage{hyperref}

\usepackage[accepted]{icml2023}

\usepackage{amsmath}
\usepackage{amssymb}
\usepackage{mathtools}
\usepackage{amsthm}

\usepackage{cleveref}

\usepackage{graphicx}
\usepackage{xcolor}
\usepackage{epigraph}

\theoremstyle{plain}

\theoremstyle{definition}

\theoremstyle{remark}

\usepackage[textsize=tiny]{todonotes}
\usepackage{multirow}

\icmltitlerunning{\textsc{Nugget}: Neural Agglomerative Embeddings of Text}

\newcommand{\nugget}[0]{\textsc{Nugget}\,}
\renewcommand{\vec}{\mathbf}
\newcommand{\nuggen}[0]{\mathtt{Nugget}}
\newcommand{\thick}{\specialrule{.1em}{.05em}{.05em}}

\begin{document}

\twocolumn[
\icmltitle{\textsc{Nugget}: Neural Agglomerative Embeddings of Text}

\icmlsetsymbol{equal}{*}

\begin{icmlauthorlist}
\icmlauthor{Guanghui Qin}{jhu}
\icmlauthor{Benjamin Van Durme}{jhu}
\end{icmlauthorlist}

\icmlaffiliation{jhu}{Department of Computer Science, Johns Hopkins University, USA}

\icmlcorrespondingauthor{Guanghui Qin}{gqin2@jhu.edu}

\icmlkeywords{natural language processing,document representation,transformers}

\vskip 0.3in
]

\printAffiliationsAndNotice{
} %

\begin{abstract}

Embedding text sequences is a widespread requirement in modern language understanding.
Existing approaches focus largely on constant-size representations.
This is problematic, as the amount of information contained in text can vary.
We propose a solution called \nugget, which encodes language into a representation based on a dynamically selected subset of input tokens.
These \emph{nuggets} are learned through tasks like autoencoding and machine translation, and intuitively segment language  into meaningful units.
We demonstrate \nugget outperforms related approaches in tasks involving semantic comparison. 
Finally, we illustrate these compact units allow for expanding the contextual window of a language model (LM), suggesting new future LMs that can condition on  larger amounts of content.
\end{abstract}

\section{Introduction}

\vspace{-4pt}

\epigraph{You can't cram the meaning of a whole  \%\&!\$\# sentence into a single \$\&!\#* vector!}{\it{Ray Mooney}}
\vspace{-5pt}

Embedding language into dense representations is a central pursuit in modern Natural Language Processing and Machine Learning. %
Recent work on  text encoding has largely focused on fixed-dimensional representations that use either one or a constant number of vectors, e.g., DAN~\citep{iyyer2015DeepUnorderedComposition}, DPR~\citep{dpr2020}, or TSDAE~\citep{tsdae2021}.
At the other extreme, {\sc ColBERT} \cite{colbert2020} represents and indexes content by storing the final {\sc BERT} \citep{bert2019} layer encoding of nearly every input token. Unfortunately a fixed dimensional representation risks not scaling to long texts, while a solution like {\sc ColBERT} comes at significant cost. We propose that a flexible balance can be found, %
leading to a \emph{``semantically useful level of granularity''}~\cite{skipprop2017}.
  
Our solution, \nugget, is
an encoding strategy  employing hard-attention to map linguistic input into a fractional number of dynamically selected embeddings called \emph{nuggets}. As the nugget selection process is non-differentiable, we build a residual connection between the selector and decoder to allow gradient propagation, enabling the model to be trained in an end-to-end manner via tasks such as autoencoding or machine translation. This approach allows the number of vectors to grow with input length, trading performance against memory as a configurable compression ratio.

\begin{figure}
    \centering
    \includegraphics[scale=0.9]{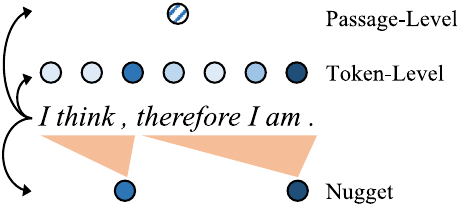}
    \caption{
    Three approaches to embedding text.
    Token-level models map each token to a vector, while passage-level models map the whole passage into a single vector.
    \nugget generates a dynamic number of vectors,
    where each nugget encodes a segment of text.
    }
    \label{fig:intro}
\end{figure}

\nugget leads to an \emph{intrinsically} interesting representation, where the encoder learns to favor clausal text delimiters, such as punctuation and conjunction words. %
Moreover, without any explicit guidance during training, each resultant nugget encodes a contiguous segment of text preceding these clausal delimiters, such as illustrated in \cref{fig:intro}. 

We demonstrate that \emph{extrinsically} these nuggets outperform prior unsupervised approaches in experiments on document-level paraphrase selection and related passage retrieval.

Finally, through an experiment on language modeling we show that \nugget can provide context information to other models in an efficient way.
Looking ahead, we believe fractional representation strategies like \nugget will allow for exciting new developments in large language models (LLMs). As nuggets support highly accurate reconstruction, they hold promise as a compressed unit of language that could  enable scaling LLMs to condition on significantly longer textual inputs.

\section{Background}

\textbf{Token-level Embeddings} are commonly used in NLP.
To map tokens to individual vectors, \citet{glove2014} uses the word co-occurrence matrix as features, while
\citet{word2vec2013} maps words to vectors by training a model to reconstruct the context.
Instead of static mappings, encoders such as CoVe~\citep{cove2017}, ELMo~\citep{elmo2018}, BERT~\citep{bert2019} and BART~\citep{bart2020} generate contextualized token embeddings.

\textbf{Unsupervised methods for passage embedding} \quad
Early related work modeled passages as topic distributions~\citep{landauer1998introduction,blei2003latent}.
With neural networks,
researchers map the sentence into one or a fixed number of vectors.
Some researchers try to derive a sentence representation from the pretrained encoder without fine-tuning \citep{wang2020SBERTWKSentenceEmbedding,li2020SentenceEmbeddingsPretrained}.
Researchers also treat it as an \emph{unsupervised learning} task.
\citet{skipthought2015} trains sentence encoding by predicting the surrounding sentences.
\citet{bowman2016GeneratingSentencesContinuous,tsdae2021,mahabadi2021VariationalInformationBottleneck} explore autoencoding to map sentences into single vectors.
With a contrastive objective, 
\citet{ct2021} learns to have similar representations of the same sentence with two independent encoders, while SimCSE \citep{simcse2021} uses different dropout masks on the same encoder.
\citet{giorgi2021DeCLUTRDeepContrastive} is similar but relies on document structure to identify positive sentence pairs.
Recently, \citet{li2022DiffusionLMImprovesControllable} propose to model texts by denoising a sequence of Gaussian vectors, leading to better controllability.

\textbf{Supervised methods for passage embedding} \quad
To construct datasets for general-purpose sentence encoders, it is common to extract sentence pairs from datasets such as natural language inference and question answering \citep{conneau2017SupervisedLearningUniversal}.
SBERT \citep{sbert2019} fine-tunes the BERT model \citep{bert2019} and uses mean pooling over the token embeddings as the sentence encoding.
In the domain of dense information retrieval,
people map documents into vectors to measure their similarity. 
Some models simply reuse the token-level encodings:
\citet{colbert2020} uses all token embeddings as the index of the document,
while \citet{dpr2020} only reuses the embedding of the \texttt{CLS} token.
\citet{gao2021CondenserPretrainingArchitecture,oguz2022DomainmatchedPretrainingTasks} show that continual training can produce information-rich \texttt{CLS} representations.

The methods mentioned above use a single vector or all tokens as the representation.
\citet{pseudoToken2022} increase the number of vectors by introducing pseudo sentences, while \citet{mvdoc2022} append \texttt{View} pseudo tokens to the BERT \cite{bert2019} self-attention; both have fixed-sized vectors, regardless of the lengths of the input. \citet{skipprop2017}, who helped inspire this work,  decomposed sentences into a variable number of \emph{propositional} embeddings, relying on a linguistic processing pipeline.

\section{Approach}

We use a modified transformer encoder-decoder architecture.
Let $\vec{w} = \{w_i\}_{i=1}^{n}$ denote the input sequence, where $n$ is the number of tokens. A transformer encoder is used to map them into contextualized embeddings:
\begin{align}
    \vec{X} = \mathtt{Encoder}(\vec{w}), \nonumber
\end{align}
where $\vec{X} \in \mathbb{R}^{n\times d}$ and $d$ is the hidden dimension.
Instead of feeding the entire $\vec{X}$ into the transformer decoder,
we use a ``nugget generator'', denoted by $\nuggen$, to produce a latent variable $\vec{Z}$ that are fed as the inputs of the decoder:
\begin{align}
    \vec{Z} &= \nuggen(\vec{X}), \nonumber \\
    p(\vec{y} \mid \vec{Z}) &= \mathtt{Decoder}(\vec{Z})
\end{align}
where $\vec{Z}\in\mathbb{R}^{k\times d}$, $k \le n$ is the number of ``nuggets'' generated by $\nuggen$, and $\vec{y}$ is the target sequence.
Note that $k$ is not a constant number and depends on $\vec{X}$.
$\mathtt{Decoder}$ is a transformer module with causal masking and is conditioned on $\vec{Z}$ via cross-attention.

In the remainder of this section
we  introduce the form of $\nuggen$ and the corresponding training strategies.

\subsection{Nugget Generator}
\label{sec:gen}

Instead of producing vectors that do not correspond to actual tokens, such as the \texttt{CLS} or averaged pooling over all token embeddings,
we leverage the fact that contextual token embeddings carry the semantics of their surrounding texts,
and use them as document representations.
We use a feedforward network to measure the amount of context information of every token embedding, then select the most informative vectors as the output:
\begin{align}
    \vec{s} &= \mathtt{FFN}(\vec{X}),
    \label{eq:scorer} \\
    \vec{X}' &= \mathtt{TopK}_k(\vec{s},\vec{X}), \label{eq:topk}\\
    \vec{Z} &= \mathtt{\nuggen} (\vec{X})  = \vec{X}'\vec{W}^V,\label{eq:opalvalue}
\end{align}
where $\vec{s}\in\mathbb{R}^n$ are a list of scores, $\mathtt{TopK}$ is an operator to pick the top $k$ elements in $\vec{X}$ sorted by $\vec{s}$, and $\vec{X'}\in\mathbb{R}^{k\times d}$ are the selected embeddings, $\vec{W}^V$ is a trainable parameter,
and $\vec{Z} \in \mathbb{R}^{k\times d}$ are the latent variables, called nuggets.

\paragraph{Choice of $k$} 
If we let $k$ be a constant, then $\nuggen$ falls back to a fixed-dimensional representation.
Instead, we let $k$ grow with the length of the text by setting $k = \lceil n\cdot r \rceil$, where the compression ratio $0<r\le1$ is a hyperparameter.

\paragraph{Alternative viewpoint} Equivalently, one can also view $\nuggen$ as \emph{hard attention}.
Let $\vec{q} \in \mathbb{R}^d$ denote a trainable query vector, and we use $\vec{X}$ as both keys and values. We can regard \cref{eq:scorer} as the attention logits:
\begin{align}
    \vec{s} = \left(\vec{q}\vec{W}^{Q}\right)\left(\vec{X}\vec{W}^{K}\right)^\top, 
    \nonumber %
\end{align}
where $\vec{W}^{Q}, \vec{W}^{K}\in \mathbb{R}^{d\times d}$ are trainable parameters.
In the next step, instead of aggregating the values $\vec{X}$, we use \emph{hard attention} to take the top-$k$ values in $\vec{X}\vec{W}^V$ with $\vec{s}$ as keys.

\begin{figure}[t]
\centering
\includegraphics[scale=0.85]{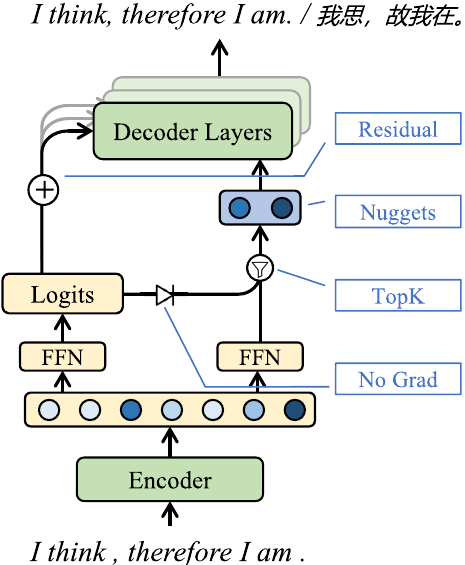}
\caption{\label{fig:forward}
The architecture of \nugget.
The diode symbol means that the gradient cannot be back-propagated.
}
\end{figure}
\subsection{Ensuring Differentiability}
\label{sec:diff}

Note that the $\mathtt{TopK}$ operator in \cref{eq:topk} is not differentiable,
thus the parameters in \cref{eq:scorer} do not receive any gradient signals.
Therefore, we build a \emph{residual connection} between the encoder and the decoder to propagate the gradients back to the $\nuggen$.
Specifically, we append the attention logits $\vec{s}$ to the cross attention in the decoder by:
\begin{align}
    \vec{a}_\iota = \frac{1}{\sqrt{d}}\left[\left(\vec{Z}\vec{W}^Q\right)\left(\vec{x}^\text{tgt}\vec{W}^K\right)^\top {\color{red} + \vec{s}} \right] , \label{eq:residual}
\end{align}
where $\vec{a}_\iota$ is the cross-attention logits for the target token $\vec{x}^\text{tgt}$ in one attention head at one of the decoder layers,
and it will be fed into a $\mathtt{SoftMax}$ operator to produce an attention distribution.
Note that we have replaced the source tokens with the nuggets $\vec{Z}$.
In addition to attending to the nugget vectors, the attention score directly takes into account the nugget logits $\vec{s}$.
As the cross-attention is differentiable, it can be viewed as a residual connection that allows the gradients to be back-propagated to the hard attention parameters.
The architecture of \nugget is shown in \cref{fig:forward}.

\textbf{Gradient analysis}\quad To interpret the gradients on $\vec{s}$, we can rewrite it as:
\begin{align}
    \frac{\partial \ell}{\partial \vec{s}} = \sum_\iota 
    \left(
        \frac{\partial \vec{a}_\iota}{\partial\vec{s}}
        \cdot
        \frac{\partial \ell}{\partial \vec{a}_\iota}\right)
    =\frac{1}{\sqrt{d}} \cdot \sum_{\iota} \frac{\partial \ell}{\partial \vec{a}_\iota},
    \label{eq:gradient}
\end{align}
where $\ell$ is the loss value, 
and the summation on the subscript $\iota$ is taken over all target tokens, attention heads, and decoder layers.
\cref{eq:gradient} shows that the gradient on the $\vec{s}$ is proportional to that on all $\vec{a}_\iota$.
Consequently, \emph{the nugget logit $s_i$ tends to increase if the model tends to pay more attention to the corresponding nugget vector $\vec{z}_i$.}
As the bottleneck of the model is to limit the number of nuggets,
the model learns to select the token embeddings that contain the maximal amount of contextual information.

Different from previous work with residual connections \citep{he2017DeepResidualLearning}, the introduction of \cref{eq:residual} to \nugget is propagating gradients to the logits $\vec{s}$,
which otherwise cannot be learned.
The absolute values of $\vec{s}$ do not greatly affect the cross-attention of the decoder,
and we do not observe much performance difference in experiments when ablating $\vec{s}$ in \cref{eq:residual} during inference.

\subsection{Informed Nugget Encoding}
\label{sec:informed}

The assumption behind \nugget is that certain tokens function as nuggets to aggregate the surrounding semantics.
However, the nugget selection is done after the encoding process,
thus cannot affect its attention behavior.
To inform the encoder of the selected nuggets, we \emph{prepone} the calculation of $\vec{s}$ to the $l$-th layer of the encoder:
\begin{align}
    \vec{s} = \mathtt{FFN}(\vec{X}^{\color{red}(l)}), \label{eq:static_logits}
\end{align}
where $\vec{X}^{(l)}$ are the hidden states of the encoder in the $l$-th layer, 
and we suppose the encoder has $L \ge l$ layers in total.
With $\vec{s}$ and the compression ratio $r$, we are able to tell apart the nugget and non-nugget tokens.
Akin to the ``segment embedding'' in \citet{bert2019}, 
we add 2 ``type embedding'' vectors, denoted by $\vec{e}^n$ and $\vec{e}^o$, to the hidden states of nugget and non-nugget tokens in the $l$-th layer, which are then fed into the next layer:
\begin{align}
    \vec{X}^{(l+1)} = \mathtt{SelfAttn}(\vec{X}^{(l)} {\color{red} + \vec{E}}), \label{eq:informed-scorer}
\end{align}
where $\vec{E}\in\mathbb{R}^{n\times d}$ are the type embedding matrix. 
We call this the \textbf{nugget feedback}.

\begin{figure}[t]
\centering
\includegraphics[scale=0.8]{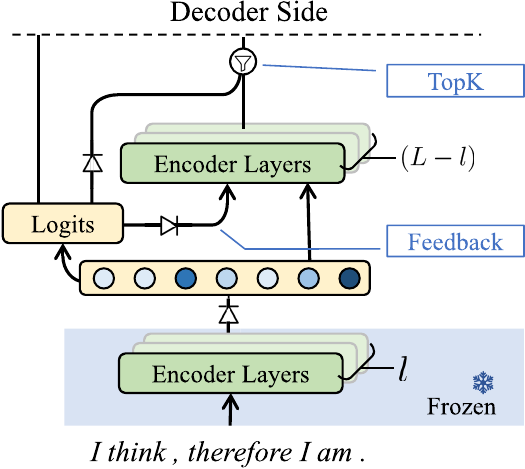}
\caption{\label{fig:informed}
The encoder of \nugget with \emph{feedback}.
The bottom $l$ layers do not receive gradient signals from back-propagation.}
\end{figure}

Note that the encoding $\vec{X}$ used in \cref{eq:topk} are still the embeddings in the last layer.
The updated nugget encoding is illustrated in \cref{fig:informed}.

\textbf{Stabilized training} \quad
In practice, we found that the training of nugget selection in \cref{eq:scorer} can be unstable when the features fed into \cref{eq:informed-scorer} are being updated.
We adopted the common practice for fine-tuning pretrained LMs~\citep{howard2018UniversalLanguageModel} to freeze the bottom $l$ layers of the encoder,
which stabilized our training curves.
\footnote{
Freezing bottom layers may also help preserve the multilingual ability of a pretrained multilingual language model; this was not tested in our experiments.
}

\subsection{Learning}

The model parameters $\theta$ are optimized by minimizing the negative log likelihood:
\begin{align}
\ell = -\sum_{\vec{w},\vec{y}\in \mathcal{D}}\log p(\vec{y}\mid \vec{w}; \theta) ,\nonumber
\end{align}
where the inputs $\vec{w}$ and outputs $\vec{y}$ are sampled from the dataset $\mathcal{D}$.
The dataset $\mathcal{D}$ can be a monolingual corpus, in which case $\vec{y}$ should be identical to $\vec{w}$ and the \nugget is trained as an \emph{autoencoder}.
Following previous work \citep{tsdae2021}, we may randomly delete tokens from $\vec{w}$ as noise.
The dataset can also be bitexts, then the target document $\vec{y}$ is translated from $\vec{w}$.
In this case, \nugget is trained as a \emph{machine translation model}~\citep{cove2017}.

\section{Experiment Setup}
\label{sec:setup}

While we could apply the \nugget concept to a variety of existing models, for experiments here we build on the architecture of BART \cite{bart2020}.  We start with the checkpoint in \citet{mbart50},
which is a model with 12 layers of encoder and decoder, and is optimized for many-to-many machine translation.
It contains 602M parameters, with 256M in the embedding matrix, 152M in the encoder and 203M in the decoder.

For the dataset, we use the English-to-Chinese subset of WMT19 corpus \citep{wmt19translate}, the same corpus used by \citet{mbart50}, as our datasets.
WMT19 is comprised of individual sentences, 
and we concatenate the adjacent sentences together to recover the document structure, similar to the practice of \citet{junczys-dowmunt2019MicrosoftTranslatorWMT}.
We limit each document to a maximum length of 128 sub-words.
The model is trained to translate English documents into Chinese documents.
For the autoencoding (AE) objective,
we use English documents on both the source and target sides.

We explored different compression ratios $r$ from 0.05 to 0.25.
We freeze the bottom 3 layers ($l=3$) in \cref{sec:informed} across our main experiments, 
and we provide a study of the effect of the number of frozen layers in \cref{sec:ablation}.
We put more training details in \cref{app:mtaetraining}.

\section{Intrinsic evaluation}
\label{sec:intrinsic}

In this section, we conduct experiments to investigate the impact of compression ratio $r$. 
We also discuss the behaviors of the nuggets and their relationship to the textual forms.

\subsection{What is a sufficient compression ratio?}
\label{sec:sufficient}

The compression ratio $r$ controls the trade-off between space efficiency and the ``semantic completeness'' of the nuggets.
Prior to applying \nugget to downstream tasks to find a sufficient compression ratio, 
we propose to use beam search with a beam size of 5 to decode texts from the generated nuggets and measure their difference from the inputs with the BLEU \citep{papineni2002BLEUMethodAutomatic} metric.

\begin{figure}
    \centering
    \includegraphics[scale=0.65]{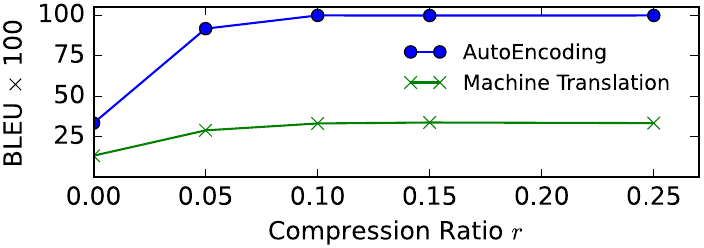}
    \caption{
    The micro-averaged BLEU value of the texts generated from nuggets with the input document as the reference.
    Note that $r=0.0$ indicates that a single vector is used for each document.
    Results are reported on the dev set of WMT19.
    }
    \label{fig:bleu}
\end{figure}

We evaluate the model on the dev set of the English-to-Chinese subset of WMT19,
where sentences are concatenated to document with a maximum length of 128 tokens.
The experiment results are shown in \cref{fig:bleu}.
With both the AE and MT training objectives, 
the performance starts to be saturated with a compression ratio of $r=0.1$.
It shows that with 10\% of tokens as nuggets,
the model has already gained sufficient information about the source documents.
In the case of autoencoding,
the BLEU value is higher than 0.99 when $r \ge 0.1$, 
meaning \nugget reconstructs the inputs nearly verbatim,
achieving almost lossless text encoding.

\subsection{What is selected as nuggets?}
\label{sec:select-analysys}

\begin{figure}
    \centering
    \includegraphics[scale=0.92]{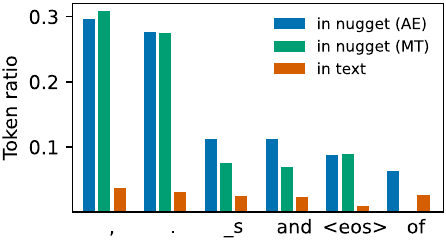}
    \caption{
    The 6 most frequent tokens selected by \nugget.
    We show their ratio in the nuggets with the AE and MT training objectives compared to that in normal texts 
    The statistics are sampled from 128 documents of lengths up to 128.
    The compression ratio is set as $r=0.1$
    for both models.
    }
    \label{fig:constituent}
\end{figure}

Instead of uniformly selecting tokens,
the scorer (\cref{eq:scorer}) of \nugget prefers certain tokens.
\cref{fig:constituent} shows the top-6 most frequent tokens selected by \nugget, and they are mostly delimiter words, like punctuation tokens (commas and periods), conjunctions, and prepositions.
Previous work on the study of transformer language models shows that a large amount of self-attention focuses on the delimiter tokens, such as punctuations, and they may be used as no-op \citet{clark2019WhatDoesBERTa}.
However, our study suggests that they may also serve as \emph{summary tokens},
as predicting the end of a segment requires the model to understand the semantics of the preceding texts.

It is worth noting that in our case study,
\nugget prefers \texttt{EOS} while \texttt{BOS} is never selected,
contrary to the practice of \citet{tsdae2021}.
Also, \nugget is not necessarily selecting the most frequent tokens.
For example: the type \emph{`the'}, which makes up 5.2\% of all tokens in the corpus, accounts for only 0.7\% of selected nuggets.
An example text is shown in \cref{fig:example}, where commas, periods, and the conjunction \emph{`and'} are selected as nuggets.
\definecolor{grey0}{RGB}{195,195,195}
\definecolor{grey1}{RGB}{182,182,182}
\definecolor{grey2}{RGB}{169,169,169}
\definecolor{grey3}{RGB}{156,156,156}
\definecolor{grey4}{RGB}{143,143,143}
\definecolor{grey5}{RGB}{130,130,130}
\definecolor{grey6}{RGB}{117,117,117}
\definecolor{grey7}{RGB}{104,104,104}
\definecolor{grey8}{RGB}{91,91,91}
\definecolor{grey9}{RGB}{78,78,78}
\definecolor{grey10}{RGB}{65,65,65}
\definecolor{grey11}{RGB}{52,52,52}
\definecolor{grey12}{RGB}{39,39,39}
\definecolor{grey13}{RGB}{26,26,26}
\definecolor{grey14}{RGB}{13,13,13}
\definecolor{grey15}{RGB}{0,0,0}
\definecolor{bgc}{RGB}{153,255,153}
\begin{figure}
\centering
\fbox{
\begin{minipage}{21em}
 {\color{grey1} Natural} {\color{grey0} language} {\color{grey2} process}{\color{grey8} ing} {\color{grey6} is} {\color{grey9} an} {\color{grey13} }{\color{grey2} interdisciplinar}{\color{grey10} y} {\color{grey8} sub}{\color{grey5} field} {\color{grey12} of} {\color{grey1} linguis}{\color{grey7} tics}\colorbox{bgc}{{\color{grey15} ,}} {\color{grey1} computer} {\color{grey5} science}\colorbox{bgc}{{\color{grey14} ,}} \colorbox{bgc}{{\color{grey14} and}} {\color{grey0} artificial} {\color{grey14} }{\color{grey5} intelligence} {\color{grey5} concerned} {\color{grey8} with} {\color{grey10} the} {\color{grey0} interaction}{\color{grey13} s} {\color{grey1} between} {\color{grey6} computer}{\color{grey13} s} \colorbox{bgc}{{\color{grey14} and}} {\color{grey1} human} {\color{grey4} language}\colorbox{bgc}{{\color{grey15} ,}} {\color{grey9} in} {\color{grey10} particular} {\color{grey5} how} {\color{grey9} to} {\color{grey9} program} {\color{grey8} computer}\colorbox{bgc}{{\color{grey14} s}} {\color{grey10} to} {\color{grey11} process} \colorbox{bgc}{{\color{grey15} and}} {\color{grey11} anal}{\color{grey12} y}{\color{grey11} ze} {\color{grey2} large} {\color{grey5} amount}{\color{grey14} s} {\color{grey13} of} {\color{grey10} natural} {\color{grey2} language} {\color{grey11} data}\colorbox{bgc}{{\color{grey15} .}} {\color{grey6} The} {\color{grey4} goal} {\color{grey8} is} {\color{grey11} a} {\color{grey7} computer} {\color{grey4} capable} {\color{grey12} of} {\color{grey9} ``}{\color{grey11} under}{\color{grey4} standing}{\color{grey12} ''} {\color{grey11} the} {\color{grey7} content}{\color{grey13} s} {\color{grey13} of} {\color{grey2} documents}\colorbox{bgc}{{\color{grey14} ,}} {\color{grey3} including} {\color{grey9} the} {\color{grey2} context}{\color{grey3} ual} {\color{grey6} nu}{\color{grey9} ances} {\color{grey13} of} {\color{grey10} the} {\color{grey3} language} {\color{grey3} within} {\color{grey8} them}\colorbox{bgc}{{\color{grey15} .}} {\color{grey5} The} {\color{grey6} technology} {\color{grey1} can} {\color{grey8} then} {\color{grey7} accurate}{\color{grey7} ly} {\color{grey2} extract} {\color{grey3} information} {\color{grey12} and} {\color{grey0} insight}{\color{grey12} s} {\color{grey2} contain}{\color{grey8} ed} {\color{grey6} in} {\color{grey5} the} {\color{grey0} documents} {\color{grey7} as} {\color{grey7} well} {\color{grey6} as} {\color{grey4} categori}{\color{grey3} ze} {\color{grey12} and} {\color{grey1} organiz}{\color{grey10} e} {\color{grey3} the} {\color{grey0} documents} {\color{grey0} themselves}\colorbox{bgc}{{\color{grey15} .}}
    \end{minipage}
    }
    \caption{
    Example texts processed by \nugget.
    Tokens in darker colors have higher scores,
    and those with green backgrounds are selected as nuggets.
    The compression ratio is set as $r=0.1$ and AE is set as the training objective.
    }
    \label{fig:example}
\end{figure}

We note that the preference of \nugget on text delimiters is not specific to English.
In \cref{app:nugget_token}, we show similar results of \cref{fig:constituent} in 9 other languages.

\subsection{What is encoded in each nugget?}
\label{sec:what_encoded}

\newcommand{\locscale}[0]{0.6}
\begin{figure}[t]
    \centering
    \includegraphics[scale=0.6]{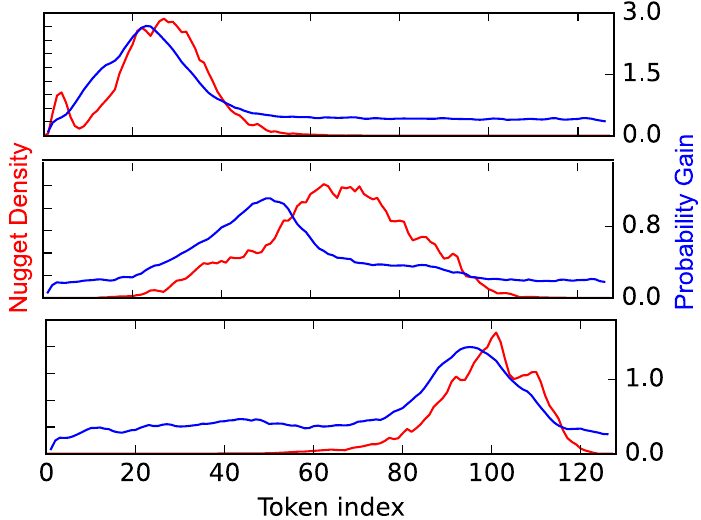}
    \caption{
    The red curve shows the distribution of token indices in the input documents of the 3rd, 6th, and 9th nuggets,
    and the blue curve shows the probability gain of every token given the corresponding nugget.
    The distribution is averaged over 10k documents.
    Compression ratio $r$ is set as $0.1$.
    }
    \label{fig:nugget-location}
\end{figure}

\begin{figure}[t]
    \centering
    \includegraphics[scale=0.58]{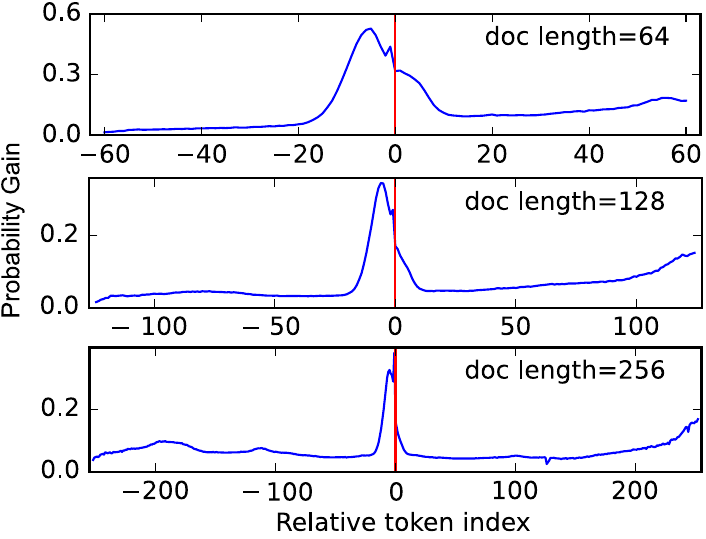}
    \caption{
    The probability gain conditioned on a single nugget.
    Graphs are averaged over all nuggets of 10k documents by centering the nugget and showing the relative indices of the tokens. The ratio $r$ is set as $0.1$.
    Refer to \cref{app:force} for a complete version.
    }
    \label{fig:nugget-rel-location}
\end{figure}

The model is optimized to encode information into nuggets, 
but it is unclear how that information is distributed across them.
Thus we propose a method to probe the semantics of individual nuggets.

We run teacher-forcing decoding on a document with a model trained with the autoencoding objective, but expose only 1 nugget during decoding.
Suppose the $j$-th nugget is exposed, then we calculate the ``probability gain'' by
\begin{align}
    g_i^j = p(y_i\mid \vec{y}_{<i}, \vec{z}_j) - p(y_i\mid \vec{y}_{<i}).
\end{align}
where $g_i^j$ measures the increment of probability mass the model has on the $i$-th token compared to the unconditional decoding. 
We order the nuggets in each document by the indices of their corresponding tokens,
and average $\vec{g}^j$ across the $j$-th nuggets of all documents.
The curves of $\vec{g}$ are plotted in \cref{fig:nugget-location}.
We can see that the exposure of a nugget can improve the decoding of its preceding texts.
Combined with our discovery in \cref{sec:select-analysys},
we speculate that \nugget is learning a \emph{divide-and-conquer} strategy,
encoding each segment with its ending delimiter tokens.

Note that this experiment made use of  documents of length 128 tokens. We then force decoded documents of lengths of 64 and 256 as well, illustrated in \cref{fig:nugget-rel-location}.
These results suggests the properties of nuggets are generalizable to documents with different lengths.

\section{What are they good for?}

With the nice properties that we observe in \cref{sec:intrinsic}, can \nugget be useful for NLP applications?
When \emph{used alone}, \nugget can be help measure the semantic similarity between texts.
\nugget can efficiently encode long texts with fewer vectors, so we  evaluate the use of \nugget in a \textbf{document similarity test}.
Also, \nugget can be used as an \emph{auxiliary module} to provide long-context semantics to other models with minimal information loss.
To focus on the language itself and exclude other factors, we propose to integrate \nugget into a language model and treat it as a \textbf{long-range sequence model}.

\subsection{Document similarity test}

It is common to use semantic textual similarity (STS) to evaluate text representation models \citep{sbert2019,tsdae2021}.
However, existing datasets for STS, such as \citet{sts2017}, are built on short sentences.
To extend this problem to long documents,
we built 2 document similarity test datasets based on the corpus of \textsc{ParaBank}~\citep{hu2019ParaBankMonolingualBitext} and WikiText-103 \citep{wikitext}.
\footnote{Those 2 datasets are released in \url{https://github.com/hiaoxui/nugget-data}}

\subsubsection{Tasks and datasets}

\paragraph{Paraphrase identification on \textsc{ParaBank}} 
\textsc{ParaBank} is a large-scale English paraphrase dataset.
It is built on single sentences that are extracted from documents, and we recover the original documents by concatenating adjacent sentences up to 256 tokens.
To make this problem difficult, sentences are randomly removed from documents and paraphrases with a probability of 20\% independently.
For each document, in addition to its paraphrase, we find another 19 negative paraphrases retrieved by the BM25 algorithm \citep{bm25},
and the model is asked to identify the correct paraphrases among 20 candidate paraphrases.

\paragraph{Passage re-ranking on WikiText-103}
WikiText-103 is a collection of Wikipedia articles.
With the leading section as the query, we randomly sample one section in the same article as the target document and retrieve 19 sections from other articles with the BM25 algorithm as negative examples.
The model is asked to rank those 20 passages according to their relevance to the leading section.

\begin{table}[t]
    \centering
    \begin{tabular}{l|lccc}
    \thick
    Task & Corpus & \#queries & $\overline{L_\text{q}}$ & $\overline{L_\text{d}}$\\
    \hline
    P\,I   & ParaBank & 1024 & 241.3 & 242.1  \\
    RR & WikiText-103 & 1024 & 287.0 & 333.8 \\
    \thick
    \end{tabular}
    \caption{
    Data statistics for the task paraphrase identification (PI) and passage re-raking (RR),
    where $\overline{L_\text{q}}$ and $\overline{L_\text{d}}$ denote the average number of tokens in query and document.
    }
    \label{tab:simdata}
\end{table}

We put the statistics of the dataset in \cref{tab:simdata}.
Please refer to \cref{app:data} for a detailed description of the dataset.

\subsubsection{Model configurations and baselines}

For those two experiments, we set the compression ratio $r$ as 0.05, 0.1, 0.15, and 0.25, and use the training objectives of both AE and MT.
We include the TSDAE model as our baseline \citep{tsdae2021}.
TSDAE is an auto-encoding model that is trained to reconstruct the input texts with the mean-pooling \footnote{
To aggregate the token embeddings, we tried using 1) mean-pooling 2) max-pooling 3) the embedding of the \texttt{CLS} token.
Consistent with the findings in table 7 in \citet{tsdae2021},
mean-pooling performs best.
}
of all the token embeddings as the bottleneck.
For fairness, we re-train the TSDAE model on WMT19 with the checkpoint of mBART under their training configurations,
where 60\%
\footnote{
We tried 0\% (no noise), but training with noise works better.
}
of input tokens are dropped as noise.
As a reference, we also tried replacing the training objective of TSDAE with machine translation.

We do not include the unsupervised models with contrastive learning objectives as baselines, such as \citet{ct2021} and \citet{simcse2021},
as they are orthogonal to our contribution: future work will consider contrastive learning for further tuning \nugget.
We refer the readers to \citet{tsdae2021} for a comparison between  contrastive learning and AE objectives.

We include the approach of ColBERT \citep{colbert2020} as a reference,
but replace the encoder with BART (that we call ``ColBART'').
ColBART uses the last hidden states of mBART encoder as the sentence embeddings.

\begin{figure}
    \centering
    \includegraphics[scale=0.7]{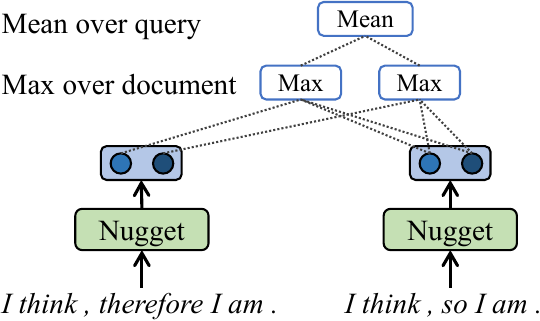}
    \caption{
    The MaxSim algorithm  in \citet{colbert2020}. 
    }
    \label{fig:maxsim}
\end{figure}

For single-vector representation models, we adopt the commonly used cosine similarity to measure the similarity between texts.
For multi-vector models (\nugget and ColBART) we adopt the \texttt{MaxSim} algorithm proposed by \citet{colbert2020} but replace the max with a mean operator because we have variable numbers of vectors:
\begin{align}
m_{q,d} = \frac{1}{I}\sum_{i} \max_j \mathtt{cos}(\vec{q}_i, \vec{d}_j), \label{eq:maxsim}
\end{align}
where $\vec{q}_i$ ($\vec{d}_j$) is the $i$-th ($j$-th) vector representation of the query $q$ (document $d$),
$I$ is the number of query vectors,
and $\mathtt{cos}$ is the cosine similarity measurement.
\footnote{We explored another 2 algorithms: 
1) Apply $\mathtt{MaxSim}$ from both sides to make it symmetric;
2) formulating it as a weighted bipartite matching problem.
We found $\mathtt{MaxSim}$ works better.}
The algorithm is illustrated in \cref{fig:maxsim}.

\subsubsection{Experiment results}

\begin{table}[t]
    \centering
    \begin{tabular}{l|ll|c|ll}
    \thick
       &ratio& obj. & multi. & P\,I & RR \\
       \hline
       \multirow{5}{*}{\rotatebox{90}{\small \nugget}} 
       & 0.25 & AE & \checkmark & 92.30 & 44.81 \\
       & 0.05 & MT & \checkmark & 92.11 &40.54 \\
       & 0.1  & MT & \checkmark & 96.69 & 50.04 \\
       & 0.15 & MT & \checkmark & 97.31 &52.36 \\
       &0.25 & MT & \checkmark & \textbf{97.38} & \textbf{56.51} \\
    \hline
    \hline
    \multicolumn{2}{l|}{\multirow{2}{*}{TSDAE}}& AE & $\times$ &  95.59 & 50.48 \\
    \multicolumn{2}{l|}{} &MT&  $\times$ & 95.04 & 45.86  \\
    \hline
    \multicolumn{3}{c|}{ColBART } & \checkmark & 94.83 & 52.44  \\
    \thick
    \end{tabular}
    \caption{
    Results on paraphrase identification (PI) and passage reranking (RR), reported as  MRR$\times$100.
    ``obj.'' denotes training objective
    and ``multi.'' denotes multi-vector representation.
    }
    \label{tab:sim}
\end{table}

Results are shown in \cref{tab:sim}.
Generally speaking, \nugget trained with the MT objective is more suitable for text similarity measurement without further tuning. 
A higher ratio leads to better performance, 
and a ratio of 0.05 (0.15) can make \nugget achieve comparable performance as ColBART does on the PI (RR) task,
while ColBART uses 20x (6.7x) more vectors to encode the text.

In practice, we found that the AE model with a low compression ratio $r$ does not perform well,
with a performance gap to TSDAE.
We speculate it is because \nugget with the AE objective does not corrupt the inputs as TSDAE does,
while \citet{tsdae2021} points out the importance of noisy training for similarity tasks.
We leave exploring noising strategies to future work.

\subsection{Long-range sequence modeling}
\label{sec:lm}

An autoregressive sequence model predicts the next token conditioned on past tokens:
\begin{align}
    p(y_i \mid \vec{y}_{1:i-1}).\label{eq:autoregressive}
\end{align}
When the contexts get longer, the computation can be costly for transformers, which suffer from their quadratic time and space complexity.
However, one can compress the history information with \nugget, and use nuggets as a substitute for the tokens.
We rewrite \cref{eq:autoregressive} as
\begin{align}
    p(y_i \mid y_{i-s:i-1}, \nuggen(\vec{y}_{1:i-s-1})),
\end{align}
where we use \nugget to encode all history tokens except for the most recent $s$ tokens.
That is, distant information is compressed before being fed into the sequence model.

\begin{figure}
    \centering
    \includegraphics[scale=0.49]{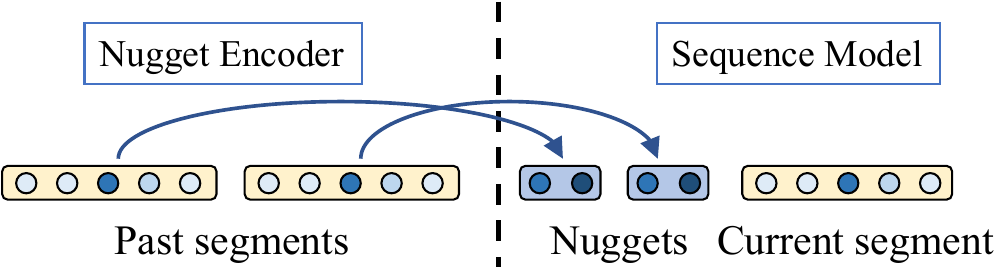}
    \caption{
    The architecture of \nugget sequence model.
    Past segments are compressed with \nugget and then fed into the sequence model, together with the tokens in the current segment.
    }
    \label{fig:lm}
\end{figure}

In experiments, we adopt the decoder part of the mBART as a language model, where the self-attention module is used to read recent tokens and the cross-attention module is used to read nuggets.
To let \nugget encoder work efficiently,
we split the distant tokens by the segment length $s$ and encode each segment independently.
The architecture of our \nugget LM is illustrated in \cref{fig:lm}.

\subsubsection{Dataset, training, and metric}

We use WikiText-103 \citep{wikitext} as the dataset with perplexity (PPL) as the evaluation metric.
Models are trained on the training set until convergence.
All the results are reported on the test set.
We exclude all out-of-vocabulary tokens during the evaluation.
\footnote{
Because mBART works on subwords with the BPE tokenizer \citep{bpe}, we take the production of the probabilities over subwords to compute the probability of the complete word.
Note that our method theoretically underestimates the model performance.
}
Please refer to \cref{app:lmtraining} for more training details.

\subsubsection{Model configurations and baselines}

We set the segment length $s$ as 128 for all the experiments.
As the context can be very long, we only encode the past $h$ segments as inputs to the language model in addition to the current segment,
where we set $h$ as 1, 2, 4, and 8,
with a corresponding context length of 256, 384, 640, and 1152.
We start \nugget LM training from the checkpoints trained with the AE objective and explored the compression ratios of 0.05 and 0.1.
As a baseline, we replace the \nugget with TSDAE and apply the same numbers of history segments.

We use compressive Transformers \citep{compressive2020} as another baseline,
which compresses the past hidden states into fewer vectors.
We adopt the ``mean pooling'' strategy in the paper and compress the most recent 512 tokens into 32 tokens, achieving a similar compression ratio as the model with $r=0.05$.
As a reference for the original transformer LM, we introduce a ``full attention model'' with a context length of 128.
It attends to all tokens without \nugget,
and is equivalent to $h=0$ in the \nugget LM experiment.

\subsubsection{Experiment results}

\begin{table}[t]
    \centering
    \begin{tabular}{l|l|cccccc}
        \thick
         \multirow{2}{*}{\rotatebox{90}{\small \textsc{Nugget}}}& & $h$=1 & $h$=2 & $h$=4 & $h$=8 \\
         \cline{2-6}
         &$r=0.05$ & 29.88 & 29.25 & 28.24 & 28.14  \\
         &$r=0.1$ & 29.83 & 29.21 & 28.44 & \textbf{28.10} \\ 
         \hline
         \multicolumn{2}{c|}{TSDAE} & 30.09 & 29.55 & 29.01 & 28.77 \\
         \hline
         \multicolumn{2}{l|}{Compressive} & - & - & 30.52 & - \\
         \hline
         \multicolumn{2}{l|}{Transformers} & \multicolumn{4}{l}{\quad($h=0$)\quad  31.46} \\
         \thick
    \end{tabular}
    \caption{
    Experiment results on language modeling.
    Performance is evaluated with perplexity (PPL).
    Above: \nugget language models with access to different numbers of history segments.
    Below: Transformer LMs with full attention with context lengths of 128 and 256, and compressive transformers \citep{compressive2020}.
    }
    \label{tab:lm}
\end{table}

Results are shown in \cref{tab:lm}.
All \nugget-assisted models can achieve lower PPL compared to full attention baseline,
meaning that the history information provided to LM is effectively utilized.
More history segments (larger $h$) are helpful,
though the improvement becomes marginal.

Though \nugget outperforms the single-vector baseline TSDAE, the difference between $r=0.05$ and $r=0.1$ is insignificant.
It might be because that $r=0.05$ has already encoded sufficient information about the content, according to our analysis in \cref{sec:sufficient}.

\section{Discussion}

\subsection{Ablation studies}
\label{sec:ablation}

As an ablation study we run \nugget without the nugget feedback (\cref{sec:informed}).
By default \nugget uses the features of layer 3 (denoted by $l=3$) to select nuggets and freeze the parameters below it.
Raising $l$ can make the features to the \nugget selector more contextualized, but also reduce the size of trainable parameters.
In the ablation study we explored setting $l$ as 0, 6, and 9,
where $l=0$ corresponds to the embedding matrix.

The selector (\cref{sec:gen})
is learned with gradient descent with the algorithm in \cref{sec:diff}.
To ablate this module
we propose 2 rule-based selectors to replace \cref{eq:scorer}:

\begin{itemize}
\item \textbf{Chunking selector}\quad  We first equally split the document into $\lceil n\cdot r\rceil$ chunks, where $n$ is the number of tokens.
For each chunk, we select the last punctuation token (comma or period) as the nugget.
If no punctuation exists in the chunk, we select the last token.
\item \textbf{Sentence boundary selector}\quad  As we concatenate the sentences in WMT19 to form documents, we use the ending tokens of sentences as the nuggets.
3.3\% of tokens are selected nuggets on average, thus we train a nugget model with $r=0.033$ as a comparison.
\end{itemize}

\begin{table}[t]
    \centering
    \begin{tabular}{l|cc}
    \thick
    Configuration & PI & RR  \\
    \hline
    \nugget ($l=3$) & \textbf{96.69} & \textbf{50.04} \\
    \quad \nugget ($l=0$) & 69.82 & 29.20 \\
    \quad \nugget ($l=6$) & 93.24 & 48.84 \\
    \quad \nugget ($l=9$) & 84.03 & 47.36 \\
    \quad No feedback & 96.29 & 49.81 \\
    \quad Chunking & 95.56 & 42.41 \\
    \hline
    \nugget ($r=0.033$) & \textbf{89.07} & \textbf{49.49} \\
    \quad Sentence boundary & 87.91 & 38.40 \\
    \thick    
    \end{tabular}
    \caption{
    The experiment results for the ablation study.
    The performance is measured by MRR$\times$100.
    }
    \label{tab:ablation}
\end{table}

We conduct experiments with those configurations on the tasks of paraphrase identification and passage reranking.
By default, we use machine translation as the training objective and use a compression ratio $r=0.1$ (or $r=0.033$ for the ``sentence boundary'' experiments).
The results are shown in \cref{tab:ablation}.
One can see that the learned nugget selector is better than rule-based selection,
and the optimal features for \cref{eq:scorer} should be derived from layer 3.
The model can also be benefited if \nugget informs the selection of nuggets via the feedback module.

\subsection{Language modeling with long contexts}

Previous work has explored ways to enlarge the effective context size for transformer-based encoders~\citep{efficient_transformers,qin2023NLPTaskEffectiveness}.
As \nugget provides certified minimal information loss with a high compression ratio, it may enable a complementary approach for long-context modeling.

Large LMs enable in-context learning (ICL) \citep{brown2020LanguageModelsAre,palm2022}, where prior task examples are concatenated as a prefix to a new example which the LM ``reasons'' over. ICL is constrained by the length of context an LM may condition on: working with compressed nuggets  may enable more ICL signal at the same context size.

\citet{wei2022ChainofThoughtPromptingElicits} demonstrated that ICL performance on complex tasks may be improved by prompting an LM to generate intermediate reasoning steps ahead of a final answer. Transformers suffer from quadratic time complexity, so decoding a \emph{chain of thought} is an expense  if one only cares about the final response.  Would it be sufficient to decode a chain of nuggets, thereby decreasing runtime?

\section{Conclusion and future work}

We proposed \nugget to encode texts with a dynamic numbers of vectors.
With auto-encoding or machine translation training, \nugget naturally segments the input texts following subsentential structures.
We demonstrate \nugget can be useful for semantic similarity and language modeling,
achieving better performance than comparable baseline models. To further improve \nugget for downstream tasks, we will consider additional training approaches such as through contrastive learning, in addition to considering applications of \nugget to large-scale language modeling.

\section*{Acknowledgement}
    
We appreciate the proofreading done by Elias Stengel-Eskin. Thanks to the anonymous reviewers for their feedback.

This research relies on the following open-source software:
PyTorch \citep{pytorch},
Lightning AI \citep{lightning},
and Huggingface Transformers \citep{huggingface_transformers}.

This work was supported by IARPA BETTER (\#2019-19051600005). 
The views and conclusions contained in this work are those of the authors and should not be interpreted as necessarily representing the official policies, either expressed or implied, or endorsements of ODNI, IARPA, or the U.S. Government. 
The U.S. Government is authorized to reproduce and distribute reprints for governmental purposes notwithstanding any copyright annotation therein.

\bibliography{ref,custom}
\bibliographystyle{icml2023}
\newpage
\appendix
\onecolumn
\section{Data construction for document similarity test}
\label{app:data}

We build two datasets for the document-level semantic similarity test.
Those 2 datasets can be downloaded in \url{https://github.com/hiaoxui/nugget-data}.
We discuss the details of the dataset construction in this section.

\subsection{Paraphrase identification}

The document-level paraphrase identification dataset is derived from \textsc{ParaBank} \citep{hu2019ParaBankMonolingualBitext}.
\textsc{ParaBank} is a large-scale English paraphrase dataset constructed with a Czech-English neural machine translation system.
We use the v1.0 of its release downloaded from \url{https://nlp.jhu.edu/parabank/}.

\textsc{ParaBank} is sentence-level, but it does not shuffle the sentence orders.
To recover the document structure, we concatenate the adjacent sentences to make ``fake documents''.
The concatenation strategy is applied to both the documents and their paraphrases by iterating their sentences in parallel
until one of them reaches the 256-token limit.
The construction process produces a list of ``\texttt{(document, paraphrase)}'' pairs.

To make the problem difficult, we delete 20\% of sentences randomly and independently on both sides.
In practice, a sentence will be included in the documents with a probability of 80\%,
and sentences are drawn independently on the document and paraphrase sides.
A robust model should be able to identify the paraphrased sentences even if they are not positionally aligned with their original sentences.

To collect negative examples, we run a BM25 algorithm \citep{bm25} with the document as the query and paraphrases as candidates.
Since the corpus \textsc{ParaBank} is too large to be efficiently indexed, and the most challenging negative examples always come from the same document, 
we try to run a sliding window around the query document with a window size of 1024 documents.
BM25 retrieves 19 negative examples from the candidates, and the model is asked to identify the correct paraphrase.

\subsection{Passage reranking}

This task asks the model to identify a document with a similar topic to the query document.
We start from the WikiText-103 data~\citep{wikitext}, which is a collection of Wikipedia articles.
We split the dataset into articles, and use the texts in sections as passages.
As the validation and test splits of WikiText are too small to generate challenging negative examples, we work on the training split.
Note that WikiText is released with a raw version and a tokenized version, and we use the raw version without masking out any \texttt{UNK} tokens.

The first section of each article is usually a general introduction about the article, thus we use it as the query document.
We randomly select another section from the same article as the answer passage,
and uses the BM25 algorithm to retrieve 19 negative examples from all but the first sections of other articles.

The statistics of the above two datasets are shown in \cref{tab:simdata}.

\section{Training details}

\subsection{Machine translation and auto-encoding training}
\label{app:mtaetraining}

We used the same codebase and training configurations for both the auto-encoding (AE) and machine translation (MT) objectives.
Both models are initialized from the checkpoint of mBART \cite{mbart50}, which is a many-to-many machine translation model.
We used the Adam \citep{kingma2015AdamMethodStochastic} optimizer with a learning rate of $5\times 10^{-5}$.
Each model is trained until convergence on the dev set.

We build a document-level MT dataset from the English-to-Chinese subset of WMT19 \citep{wmt19translate}.
The dataset is constructed so that adjacent sentences are concatenated to make document \citep{junczys-dowmunt2019MicrosoftTranslatorWMT} with up to 128 tokens.
The document might not be full and always contain complete sentences, as we do not break the sentences.
The MT model is trained to translate English into Chinese.
For the AE objective, we use the same dataset but replace the target Chinese documents with the inputs.

Every model is trained on 4 NVIDIA RTX 6000 GPUs with 24GB $\times$ 4 GPU memory.
With a batch size of 16 on each card,
the MT model can converge in approximately 48 hours.
The AE model usually converges in 36 hours.

\subsection{Language model training}
\label{app:lmtraining}

To be fair, each language model is initialized from a checkpoint of the AE model,
even if they do not require the input of history nuggets.
In practice, the transformer and the compressive transformer baselines are initialized from the AE model with $r=0.1$.
Thus, all models have the same number of parameters in the self-attention module,
while the baseline models do not utilize the cross-attention part.

The WikiText-103 data are segmented into chunks of 128 tokens, and the model is trained to predict each segment based on a certain amount of history information.
Note that during training, the training loss is calculated for all tokens in a segment in parallel,
while during inference we input the model with as many preceding tokens as possible in the current segment to provide sufficient context, up to 128 tokens.

All models are trained with 4 NVIDIA RTX 6000 GPU cards with 24$\times$4 GB GPU memories.
Adam \citep{kingma2015AdamMethodStochastic} is used and is configured with a learning rate of $5\times 10^{-5}$.
It takes around 48 hours for a model with nuggets to converge.
The model without nuggets, including the TSDAE baseline, can be faster to converge, taking around 24 hours.

\section{Analysis of \nugget encoding: Complete results}
\label{app:force}

\begin{figure}
    \centering

    \begin{subfigure}{0.48\textwidth}
         \centering
         \includegraphics[scale=0.5]{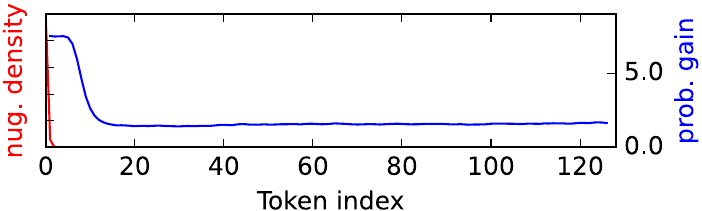}   
    \end{subfigure}
    \begin{subfigure}{0.48\textwidth}
         \centering
         \includegraphics[scale=0.5]{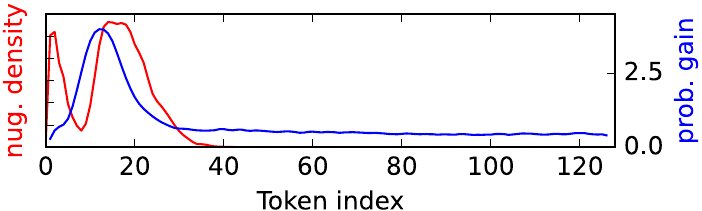}   
    \end{subfigure}
    \begin{subfigure}{0.48\textwidth}
         \centering
         \includegraphics[scale=0.5]{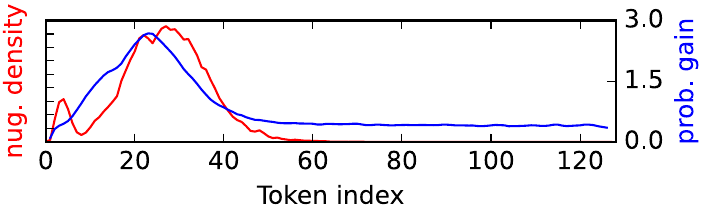}   
    \end{subfigure}
    \begin{subfigure}{0.48\textwidth}
         \centering
         \includegraphics[scale=0.5]{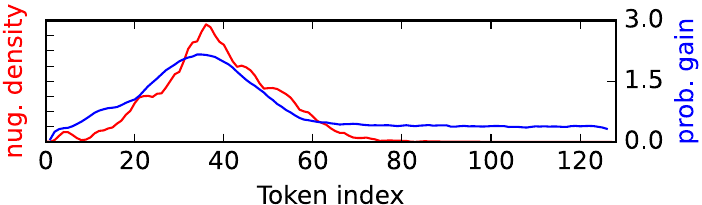}   
    \end{subfigure}
    \begin{subfigure}{0.48\textwidth}
         \centering
         \includegraphics[scale=0.5]{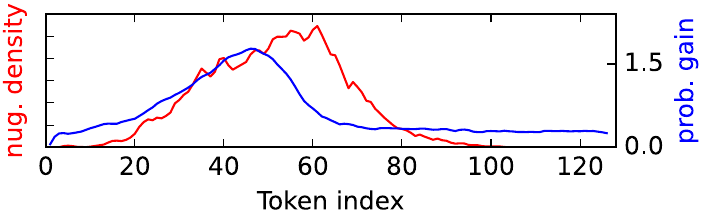}   
    \end{subfigure}
    \begin{subfigure}{0.48\textwidth}
         \centering
         \includegraphics[scale=0.5]{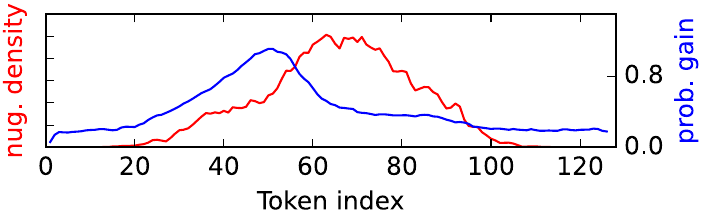}   
    \end{subfigure}
    \begin{subfigure}{0.48\textwidth}
         \centering
         \includegraphics[scale=0.5]{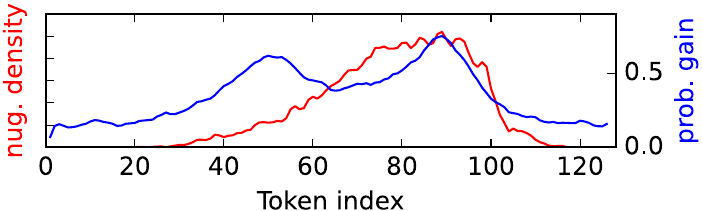}   
    \end{subfigure}
    \begin{subfigure}{0.48\textwidth}
         \centering
         \includegraphics[scale=0.5]{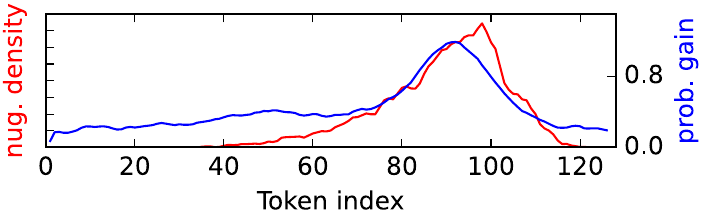}   
    \end{subfigure}
    \begin{subfigure}{0.48\textwidth}
         \centering
         \includegraphics[scale=0.5]{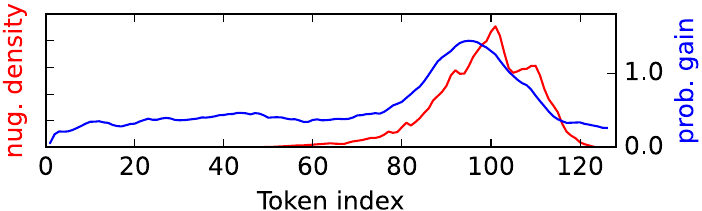}   
    \end{subfigure}
    \begin{subfigure}{0.48\textwidth}
         \centering
         \includegraphics[scale=0.5]{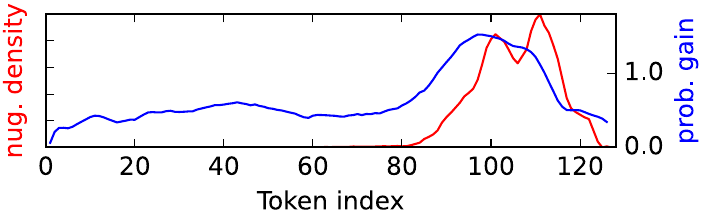}   
    \end{subfigure}
    \begin{subfigure}{0.48\textwidth}
         \centering
         \includegraphics[scale=0.5]{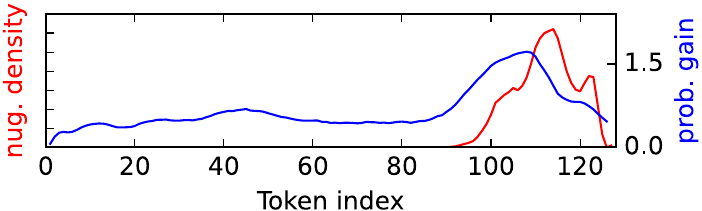}   
    \end{subfigure}
    \begin{subfigure}{0.48\textwidth}
         \centering
         \includegraphics[scale=0.5]{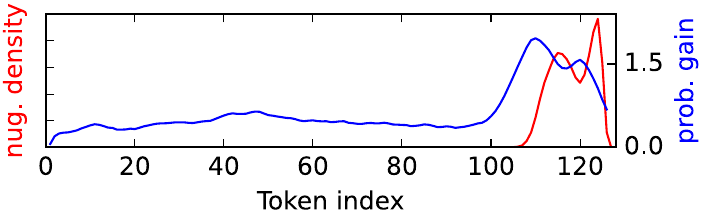}   
    \end{subfigure}
    \begin{subfigure}{0.48\textwidth}
         \centering
         \includegraphics[scale=0.5]{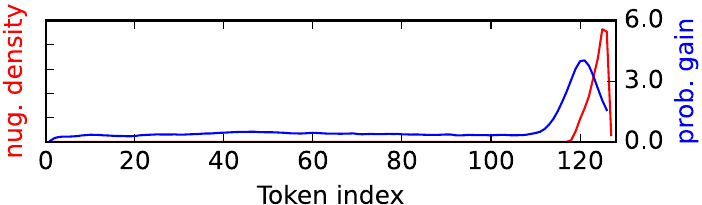}   
    \end{subfigure}
    \begin{subfigure}{0.48\textwidth}
         \centering
         \quad
    \end{subfigure}
    
    \caption{
    The probability gain on individual tokens vs the nugget location.
    We use the same notation as that in \cref{fig:nugget-location}.
    Each graph corresponds to one nugget in the texts,
    where nuggets are ordered by their indices on the original documents.
    Results are averaged over 10k documents, and only the first 13 nuggets in each document are considered.
    }
    \label{fig:force}
\end{figure}

In this section, we show a complete version of \cref{fig:nugget-location}. 
We collect the first 13 nuggets in each document.
Results are shown in \cref{fig:force}.
Please refer to \cref{sec:what_encoded} for a description of the experiments.

\section{Nugget token distribution in languages other than English}
\label{app:nugget_token}

In this section, we show the results of \cref{fig:constituent} in languages other than English.
We use all of the 9 languages from WMT19\,\citep{wmt19translate}:
Chinese, Czech, Finnish, French, German, Gujarati,  Kazakh, Lithuanian, and Russian.
Except that French is translated into German, other languages are all translated into English when the training objectives are set as machine translation.
Note that Kazakh and Gujarati have much less training data than other languages and the training on them quickly stops.
Results are shown in \cref{fig:all_constituent}.
Please refer to \cref{sec:select-analysys} for a description of the experiments.

\newcommand{\conscale}{0.6}
\newcommand{\conwidth}{0.32}
\begin{figure}
    \centering

    \begin{subfigure}{\conwidth\textwidth}
         \centering
         \includegraphics[scale=\conscale]{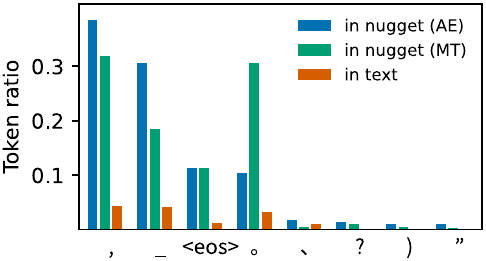}
         \subcaption{Chinese}
    \end{subfigure}
    \begin{subfigure}{\conwidth\textwidth}
         \centering
         \includegraphics[scale=\conscale]{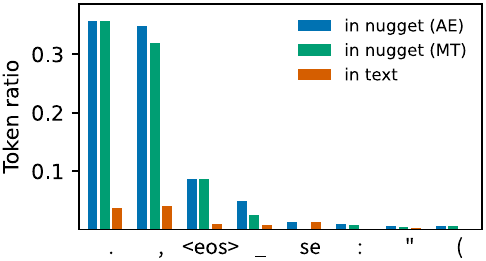} 
         \subcaption{Czech}  
    \end{subfigure}
    \begin{subfigure}{\conwidth\textwidth}
         \centering
         \includegraphics[scale=\conscale]{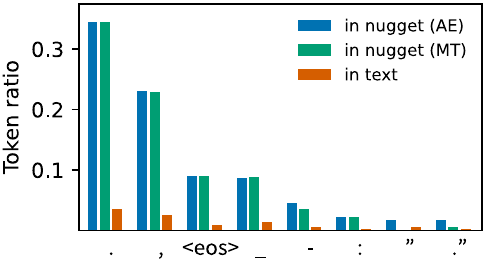}
         \subcaption{Finnish}     
    \end{subfigure}
    \begin{subfigure}{\conwidth\textwidth}
         \centering
         \includegraphics[scale=\conscale]{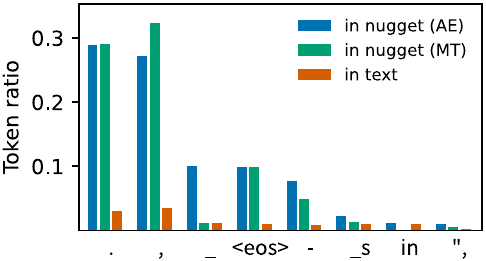} 
         \subcaption{French}       
    \end{subfigure}
    \begin{subfigure}{\conwidth\textwidth}
         \centering
         \includegraphics[scale=\conscale]{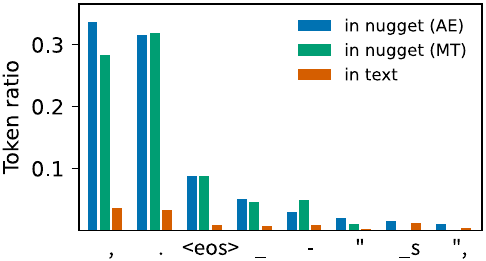}   
         \subcaption{German}        
    \end{subfigure}
    \begin{subfigure}{\conwidth\textwidth}
         \centering
         \includegraphics[scale=\conscale]{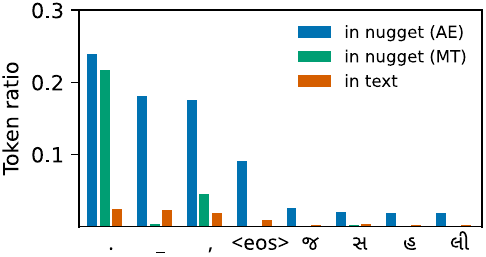}   
         \subcaption{Gujarati}     
    \end{subfigure}
    \begin{subfigure}{\conwidth\textwidth}
         \centering
         \includegraphics[scale=\conscale]{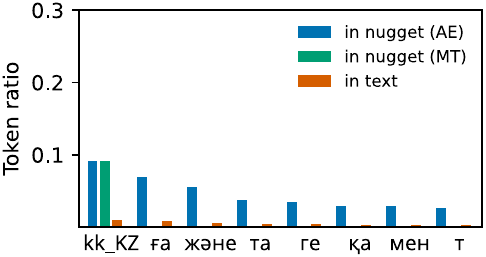}   
         \subcaption{Kazakh}     
    \end{subfigure}
    \begin{subfigure}{\conwidth\textwidth}
         \centering
         \includegraphics[scale=\conscale]{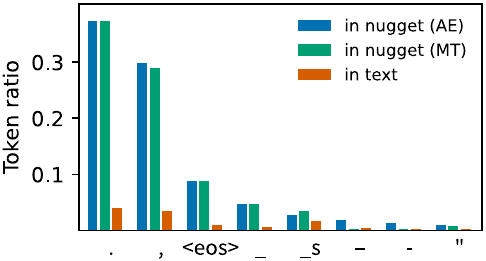}   
         \subcaption{Lithuanian}     
    \end{subfigure}
    \begin{subfigure}{\conwidth\textwidth}
         \centering
         \includegraphics[scale=\conscale]{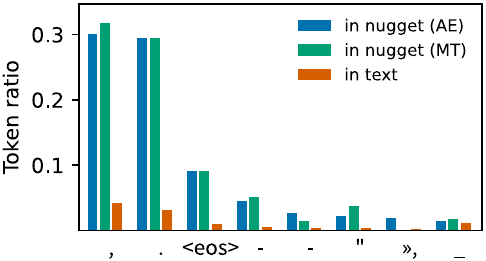}  
         \subcaption{Russian}        
    \end{subfigure}
    \caption{
    The token frequency in text and nuggets with training objectives of autoencoding (AE) and machine translation (MT).
    The experiments inherit the settings in \cref{fig:constituent} and \cref{sec:select-analysys} and are conducted in 9 other languages.
    }
    \label{fig:all_constituent}
\end{figure}
\end{document}